\documentclass[runningheads]{llncs}

\usepackage{eccv}

\usepackage{eccvabbrv}

\usepackage{graphicx}
\usepackage{booktabs}

\usepackage{amsmath,amsfonts}
\usepackage{algorithmic}
\usepackage[linesnumbered,ruled,vlined]{algorithm2e}

\usepackage{multirow}
\usepackage{threeparttable}
\newcommand{\myPara}[1]{\vspace{.05in}\noindent\textbf{#1}}

\usepackage[accsupp]{axessibility}  

\usepackage{hyperref}

\usepackage{orcidlink}

\begin{document}

\title{Learning Differentially Private Diffusion Models via Stochastic Adversarial Distillation} 

\titlerunning{Learning DPDMs via Stochastic Adversarial Distillation}

\author{Bochao Liu\inst{1,2} \and
Pengju Wang\inst{1,2} \and
Shiming Ge\inst{1,2}}\footnotetext{Shiming Ge is the corresponding author (geshiming@iie.ac.cn).}

\authorrunning{B.~Liu et al.}

\institute{Institute of Information Engineering, Chinese Academy of Sciences, Beijing\\ 100085, China \and
School of Cyber Security, University of Chinese Academy of Sciences, Beijing 100049, China}

\maketitle

\begin{abstract}
  \label{sec:abs}
  While the success of deep learning relies on large amounts of training datasets, data is often limited in privacy-sensitive domains. To address this challenge, generative model learning with differential privacy has emerged as a solution to train private generative models for desensitized data generation. However, the quality of the images generated by existing methods is limited due to the complexity of modeling data distribution. We build on the success of diffusion models and introduce DP-SAD, which trains a private diffusion model by a stochastic adversarial distillation method. Specifically, we first train a diffusion model as a teacher and then train a student by distillation, in which we achieve differential privacy by adding noise to the gradients from other models to the student. For better generation quality, we introduce a discriminator to distinguish whether an image is from the teacher or the student, which forms the adversarial training. Extensive experiments and analysis clearly demonstrate the effectiveness of our proposed method.
\end{abstract}

\section{Introduction}
\label{sec:intro}
Data sharing is essential for the development of deep learning, especially computer vision. However, in many application contexts~\cite{ge2017detecting,ge2018low,he2019part}, the sharing of data is restricted owing to its confidential nature (such as personal information on mobile devices, medical records, and financial transactions) along with strict regulatory requirements, thereby substantially impeding the advancement of technology. Data generation with differential privacy~(DP)~\cite{dwork2006calibrating,dwork2014algorithmic} can be a solution for data release without compromising privacy, where only a sanitized form of the data is publicly released. This sanitized synthetic data can be used as a substitute for actual data, analyzed using standard toolchains, and openly shared with the public, promoting technological progress and reproducible research in areas involving sensitive information.

Existing differentially private generative methods mainly focus on developing privacy-preserving generative adversarial networks (GANs), as initially introduced by~\cite{goodfellow2014generative}. They typically employ either differentially private stochastic gradient descent (DPSGD)~\cite{abadi2016deep}, or the private aggregation of teacher ensembles (PATE)~\cite{papernot2016semi}. DPSGD-based methods~\cite{cao2021don,xie2018differentially,chen2020gs,Dockhorn2022DifferentiallyPD} achieved DP by perturbing the gradients in each iteration and PATE-based methods~\cite{wang2021ccs,long2021g,jordon2018pate} achieved DP by aggregating noise labels from teachers. These methods provided an alternative to direct data release by releasing well-trained generative models that users can use to generate data for their own downstream tasks. 

However, generating high-utility data while ensuring differential privacy guarantees presents a significant challenge. There are three main shortcomings: (i) GANs are known to be considerably difficult to train, which becomes even harder when considering the privacy constraints; (ii) As the dimensionality of the data or the network escalates, an augmented quantity of noise is necessitated to attain an equivalent degree of privacy, potentially engendering more pronounced declines in performance; (iii) adding DP noise directly to all gradients introduces too much randomness, which causes damage to the quality of the generated data.

With the advent of diffusion models~\cite{ho2020denoising}, some works~\cite{Dockhorn2022DifferentiallyPD,Saiyue2023dpldm,ghalebikesabi2023differentially} wanted to address the above shortcomings by training privacy-preserving diffusion models. Despite some achievements, it leads to new problems, where training diffusion models with differentially private algorithms~(e.g. DPSGD) directly leads to excessive privacy consumption and requires pre-training on large datasets.

\begin{figure}[!t]
    \centering
    \includegraphics[width=1\linewidth]{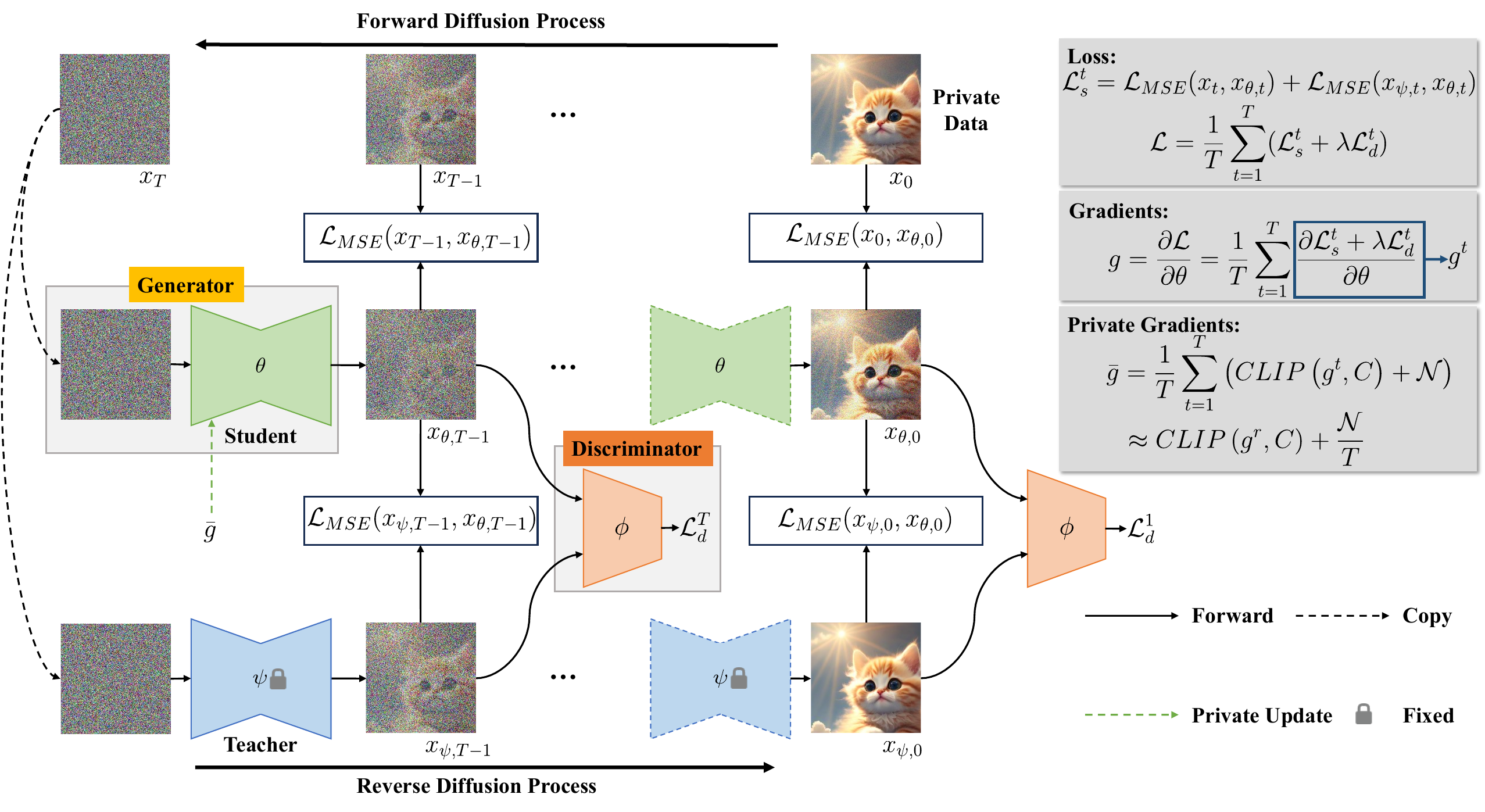}
    \caption{Overview of our DP-SAD. We first train a teacher model $\psi$ using the private data without protection. After that, we train a student model with the private data and the fixed teacher model in a distillation manner. In addition, we add a discriminator and view the student model as a generator to form adversarial training. Finally, for privacy, we achieve differential privacy by clipping with bound $C$ and adding noise $\mathcal{N}$ to the gradients during backpropagation. Furthermore, we accelerate the training by using the gradient of a random time step $CLIP(g^r,C)$ instead of averaging the gradients of all time steps in the reaction process $\frac{1}{T}\sum_{t=1}^TCLIP(g^t,C)$, where $CLIP(*,C)=*/max(1,\frac{||*||_2}{C})$.}
    \label{fig:framwork}
\end{figure}

In this work, we abandon GANs and train a privacy-preserving diffusion model with a stochastic adversarial distillation method. As shown in Fig.~\ref{fig:framwork}, in contrast to existing methods, we cleverly utilize the time step of the diffusion models to dilute the effect of DP noise and combine diffusion distillation to obtain a more stable training process. Moreover, we add a discriminator to determine whether an image is generated by the teacher or the student, which can accelerate the convergence process while enhancing the quality of the data generated by the model. As an added benefit, in contrast to other DPSGD-based methods that require a large batch size to minimize the effect of DP noise, our method can take a smaller batch size and a larger time step to achieve the same effect, which allows our method to be trained in resource-constrained scenarios.

In conclusion, our DP-SAD adeptly generates privacy-preserving images by incorporating three principal components. Firstly, it employs the time step of diffusion models to reduce the impact of DP noise, while maintaining privacy without detriment to image quality. Secondly, the introduction of a discriminator facilitates adversarial training with the student model, thereby augmenting the student model's performance. Lastly, by invoking the chain rule of gradients and capitalizing on the post-processing property of differential privacy, our method effectively minimizes the introduction of randomness. These strategic implementations collectively ensure the generation of high-utility, privacy-preserving images, underscoring the efficacy and innovation of our DP-SAD.

We summarize our main contributions as follows: i) we propose a differentially private generative modeling framework named DP-SAD for effective privacy-preserving data generation; ii) we cleverly utilize the properties of the diffusion models to reduce the impact of DP noise. Combined with a discriminator, efficient and high-performance model training is achieved while allowing for resource-constrained training as an added benefit; iii) we conduct extensive experiments and analysis to demonstrate the effectiveness of our method.

\section{Related Works}

\subsection{Diffusion Distillation.} In the advancing field of diffusion models, distillation has become a key method for enhancing model efficiency and deployment on resource-constrained platforms. Several notable works have contributed to this area by focusing on different aspects of distillation and application. Google's works~\cite{salimans2022progressive,zhao2023mobilediffusion,li2024snapfusion} have significantly pushed the boundaries in rapid sampling and mobile device applicability. These studies highlight the potential for real-time, high-quality generative tasks on handheld devices. Further, \cite{meng2023distillation,sauer2023adversarial} explored the optimization of guidance mechanisms and adversarial training in the distillation process, offering insights into the refinement of model efficiency and robustness. Innovations in data-free distillation and quality enhancement are showcased in works like \cite{gu2023boot,luo2023latent}, which proposed novel methods for minimizing dependency on large datasets and improving image resolution, respectively. Additionally, \cite{katzir2023noise,kodaira2023streamdiffusion} presented methods for reducing noise in the distillation process and enabling real-time interactive generation, highlighting the diversity of challenges and solutions in the diffusion model ecosystem. These efficient, high-quality diffusion distillation methods inspired our work, which, to our best knowledge, is the first to train a privacy-preserving diffusion model without pertaining.

\subsection{Differentially Private Generative Models.}
Training a DP generative model is a popular solution to the problem of privacy leakage in data sharing. Existing methods typically adopt DPSGD~\cite{xie2018differentially,chen2020gs,cao2021don,Dockhorn2022DifferentiallyPD,ghalebikesabi2023differentially,Saiyue2023dpldm} or PATE~\cite{wang2021ccs,long2021g,jordon2018pate} equip the generative models with rigorous privacy guarantees. These methods, despite significant breakthroughs in the training stability problem and the visual quality problem, are far from the data utility of standardized. This is because the effect of differential privacy noise is not well minimized. Utilizing the post-processing of DP does reduce the number of additions to the noise, but does not inherently reduce its effect on the gradient. In our work, we cleverly utilize the time step of the diffusion model to dilute the effect of DP noise to improve model performance and training stability.

\section{Background}

\subsection{Denoising Diffusion Probabilistic Models}
Denoising diffusion probabilistic models~\cite{ho2020denoising} are recently emerged generative models that have achieved state-of-the-art results across diverse computer vision problems~\cite{blattmann2023stable,wang2024sam}. It contains both forward and reverse processes. The forward process is a Markov chain that sequentially adds noise to a real data sample $x_0$ to obtain a pure noise distribution $x_T$, which can be understood as a labeling process. The reverse process learns the noise labels for each step in the forward process with a deep neural network to denoise $x_T$ back to $x_0$.

Given a real data sample $x_0$, we define a posterior probability according to a variance schedule $\alpha_{[1...T]}$ as follows,
\begin{equation}\label{eq:posterior}
    \begin{aligned}
        q(x_t|x_{t-1}) =\mathcal{N}(x_t;\sqrt{\alpha_t}x_{t-1}, 1-\alpha_t \textbf{I}),
    \end{aligned}
\end{equation}
where $\mathcal{N}(x;\mu, \sigma^2)$ represents $x$ obeys a Gaussian distribution with $\mu$ as the mean and $\sigma^2$ as the variance. The reverse process is parameterized by a deep neural network $\epsilon_\theta(x_t,t)$ which predicts the noise $\epsilon$ added in the forward process at step $t$. So a simplified training loss to learn $\theta$ is as follows,
\begin{equation}\label{eq:diffusion-loss}
    \begin{aligned}
        \mathcal{L}(\theta) = \mathbb{E}_{x_0,t}[||\epsilon-\epsilon_\theta(x_t,t)||^2],
    \end{aligned}
\end{equation}
where $x_t = \sqrt{\prod_i\alpha_i}x_0 + \sqrt{1-\prod_i\alpha_i}\epsilon$. In inference time, model $\epsilon_\theta(\cdot)$ can denoise a pure noise distribution to a realistic image.

\subsection{Differential Privacy}
Differential privacy is currently an industry standard of privacy proposed by~\cite{dwork2006calibrating,dwork2014algorithmic}. It limits the extent to which the output distribution of a randomized algorithm changes in response to input changes. The following definition describes how DP provides rigorous privacy guarantees clearly.
\begin{definition}[Differential Privacy]\label{def:dp}
\itshape{A randomized mechanism $\mathcal{A}$ with domain $\mathcal{R}$ is $(\varepsilon, \delta)$-differential privacy, if for all $\mathcal{O} \subseteq \mathcal{R}$ and any adjacent datasets $\mathcal{D}$ and $\mathcal{D}^{\prime}$ :}
\begin{equation}
    \begin{aligned}
    Pr[\mathcal{A}(\mathcal{D}) \in \mathcal{O}] \leq e^\varepsilon \cdot Pr\left[\mathcal{A}\left(\mathcal{D}^{\prime}\right) \in \mathcal{O}\right]+\delta,
    \end{aligned}
\end{equation}
\end{definition} 
where adjacent datasets $\mathcal{D}$ and $\mathcal{D}^{\prime}$ differ from each other with only one training example. $\varepsilon$ is the privacy budget, which measures the degree of privacy protection of the algorithm, with smaller representing better privacy protection, and $\delta$ represents the failure probability of the algorithm, which is usually set to $10^{-5}$. 

Post-processing~\cite{dwork2014algorithmic} is an important nature for privacy protection, which is described as follows:
\begin{theorem}[Post-processing]\label{th:post-processing}
\itshape{If mechanism $\mathcal{A}$ satisfies $(\varepsilon, \delta)$-DP, the composition of a data-independent function $\mathcal{F}$ with $\mathcal{A}$ also satisfies $(\varepsilon, \delta)$-DP.}
\end{theorem}   

\section{Method}

\subsection{Problem Formulation}
Given a dataset $\mathcal{D}=\{x_i\}_{i=1}^n$, the objective is to train a privacy-preserving generative model $\epsilon_\theta$ with parameter $\theta$ for high-utility data generation. To achieve this, we introduce a differentially private generative modeling method named DP-SAD, which contains three parts: teacher model $\epsilon_\psi$, student model $\epsilon_\theta$ and discriminator $\epsilon_\phi$. The training process can be formulated by minimizing an energy function $\mathbb{E}$ as follows,
\begin{equation}
    \begin{aligned}
        \mathbb{E}(\epsilon_\theta;\mathcal{D})&=\mathbb{E}_t(\epsilon_\psi;\mathcal{D}) + \mathbb{E}_{s}(\epsilon_\theta,\epsilon_\phi;\epsilon_\psi)\\
        &=\mathbb{E}_t(\epsilon_\psi;\mathcal{D}) + 
        \mathbb{E}_a(\epsilon_\phi;\epsilon_\psi,\epsilon_\theta) +\mathbb{E}_d(\epsilon_\theta;\epsilon_\psi,\epsilon_\phi),
    \end{aligned}
\end{equation}
where teacher energy $\mathbb{E}_t$ and student energy $\mathbb{E}_s$ are used to evaluate knowledge extraction and knowledge transfer respectively. We solve it via three steps: teacher learning to achieve $\epsilon_\psi$, adversarial learning to get $\epsilon_\phi$ and stochastic step learning to transfer knowledge from $\epsilon_\psi$ to $\epsilon_\theta$. We emphasize that, unlike~\cite{ho2020denoising} where the model predicts noise added in the forward process, in this paper, models~($\epsilon_\psi, \epsilon_\theta$) predict the image of next time step, and the two are equivalent.

\subsection{Teacher Learning}
We first train a teacher model using private data without any protection. This model is only used in the student training process to guide the student and is not released. We follow the standard classifier-free diffusion guidance method~\cite{ho2022classifier} to solve the energy $\mathbb{E}_t(\epsilon_\psi;\mathcal{D})$. 
\begin{equation}
    \begin{aligned}
        x_t = \mathcal{S}((1+w)\epsilon_\psi(x_{t-1}, y)-w\epsilon_\psi(x_{t-1})),
    \end{aligned}
\end{equation}
where $\mathcal{S}$ is a sampler function, $w$ is a hyperparameter and $y$ is the label of $x_{t-1}$. In this way, we obtain almost the same performance as the classifier-guided diffusion model without the need for a classifier. For datasets that are either unlabeled or multi-labeled, we employ unsupervised classification methods (e.g., k-means~\cite{macqueen1967some}) to assign labels. In our experiments, we first utilize MoCo~\cite{chen2020improved} for feature extraction, followed by the application of k-means for clustering.

\begin{figure}[t]
	\centering
	\includegraphics[width=0.49\linewidth]{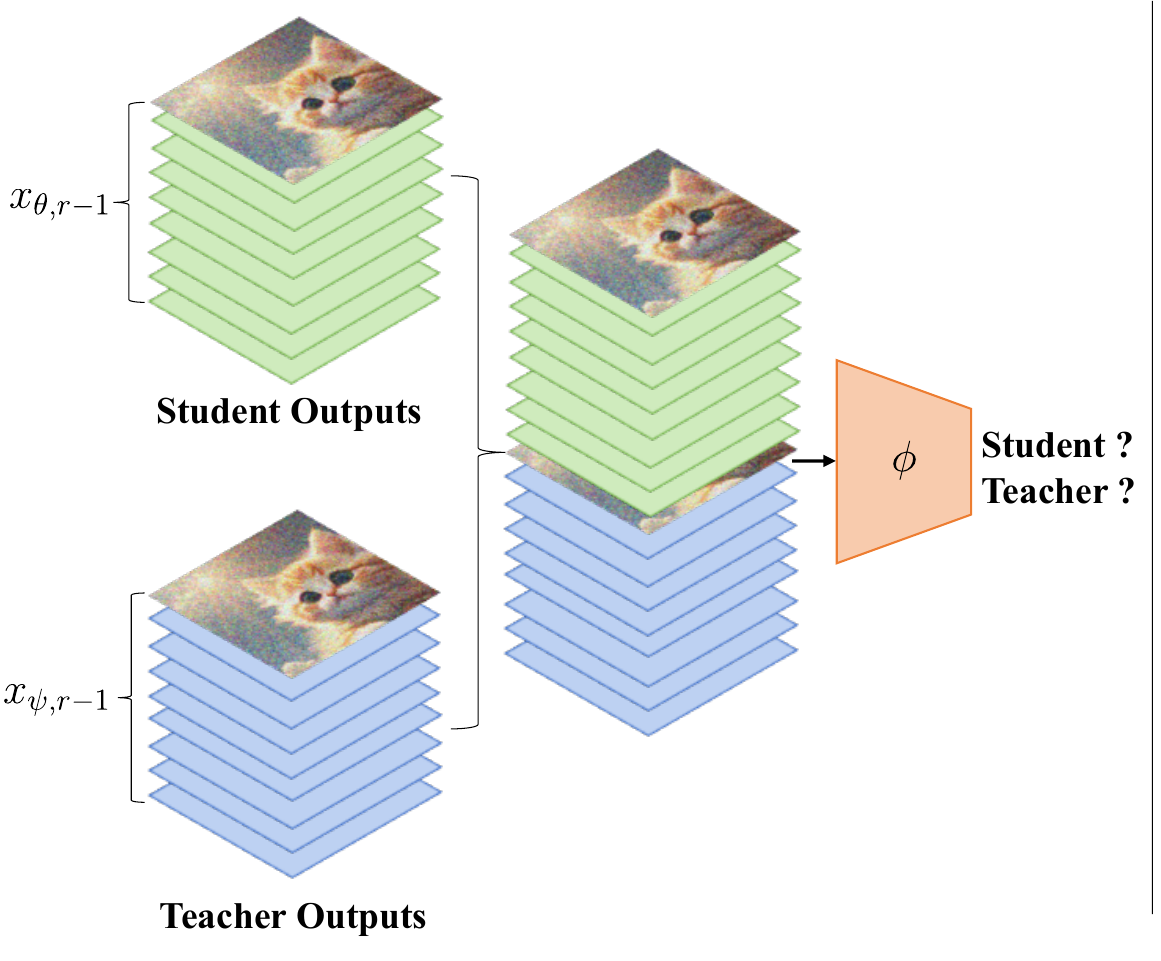}
	\includegraphics[width=0.48\linewidth]{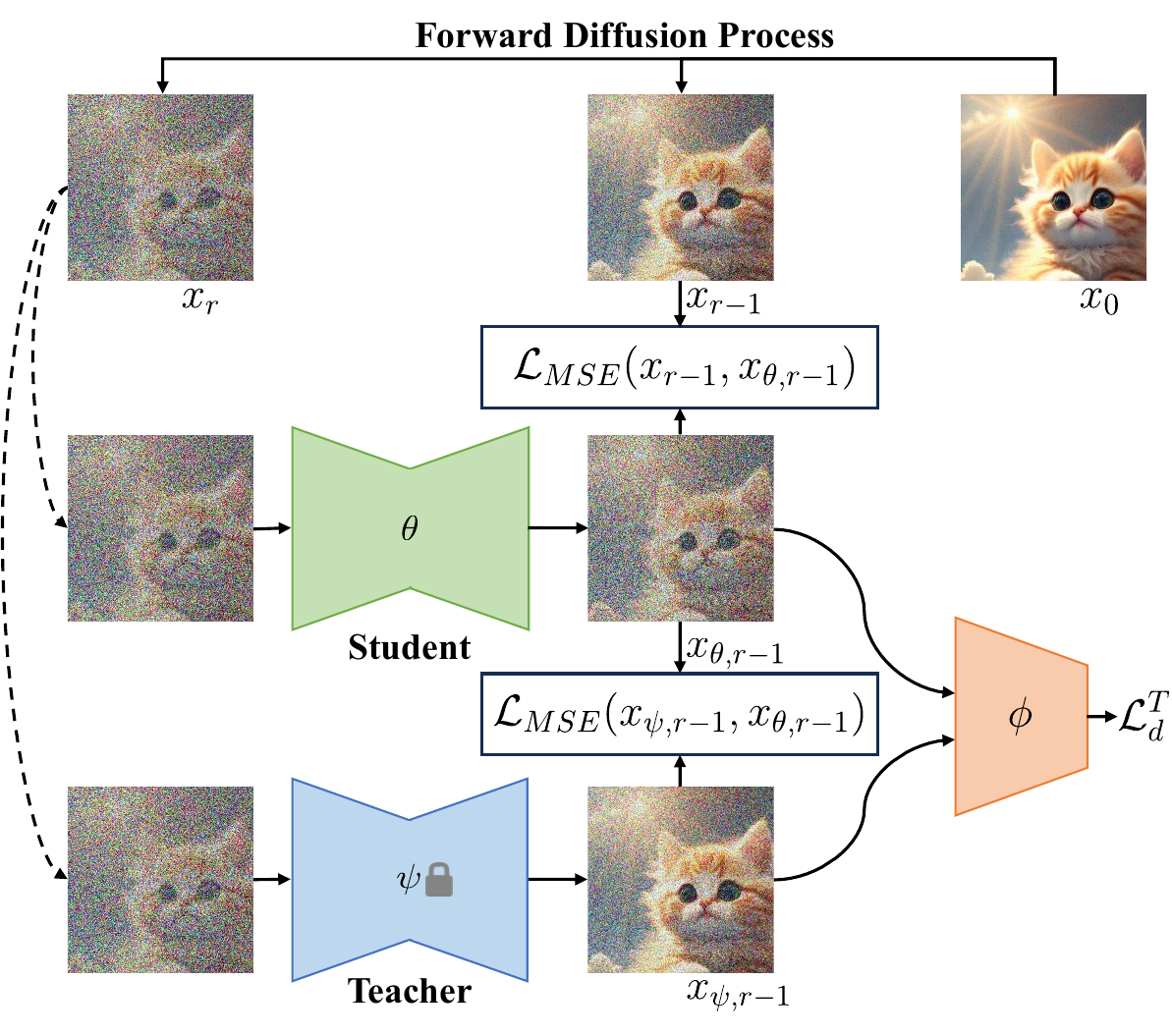}
	\caption{Left: Illustration of the proposed discriminator. We concatenate the outputs of the teacher and student models, utilizing these combined outputs as the inputs for the discriminator. The discriminator distinguishes whether the input image originates from the teacher model or the student model. Right: Use the gradients of step $r$ as a substitute for the average gradients over $T$ steps. By obtaining $x_r$ through the forward process, we prevent the teacher from inferring from noise to $x_r$, thereby saving computational time.}
	\label{fig:loss}
\end{figure}

\subsection{Adversarial Learning}
We treat the student model $\epsilon_\theta$ as a generator and introduce a discriminator $\epsilon_\phi$ to form adversarial training. The discriminator endeavors to categorize its inputs as either originating from the teacher or the student model by minimizing the following objective function~\cite{goodfellow2014generative}:
\begin{equation}
    \begin{aligned}
        \mathcal{L}_{adv}^{i,t}=\log \epsilon_\phi(x_{\psi,i,t-1}) + \log (1-\epsilon_\phi(x_{\theta,i,t-1})),
    \end{aligned}
\end{equation}
where $x_{\psi,i,t-1}$ and $x_{\theta,i,t-1}$ correspond to the $i-th$ outputs of the teacher and student models at time step $t-1$, respectively. Simultaneously, the student model aims to produce outputs closely resembling those of the teacher model, to deceive the discriminator, by minimizing the loss function $\mathcal{L}_{adv}^{i,t}$.

Since our teacher model is fixed, the first term in the loss function can be removed when updating the student model. Therefore, our loss function can be simplified as follows:
\begin{equation}\label{eq:loss_d}
    \begin{aligned}
        \mathcal{L}_{adv}^{i,t}=\log (1-\epsilon_\phi(x_{\theta,i,t-1})).
    \end{aligned}
\end{equation}
To maintain the same format as Eq~(\ref{eq:loss_d}) when updating the discriminator, we concatenate the outputs of the teacher and student models together as the input for the discriminator as shown in the left of Fig.~\ref{fig:loss}. The output corresponding to the teacher model is labeled as $[1,0]$, while the output from the student model is labeled as $[0,1]$. So the adversarial loss can be formulated as follows:
\begin{equation}
    \begin{aligned}
        \mathcal{L}_{adv}^{i,t}=\log (1-\epsilon_\phi(\mathcal{C}(x_{\psi,i,t-1},x_{\theta,i,t-1}))).
    \end{aligned}
\end{equation}
where $\mathcal{C}$ represents the concatenation function. The loss function~($\mathbb{E}_a(\epsilon_\phi;\epsilon_\psi,\epsilon_\theta)$) for a batch of data over the entire time steps $T$ is given as:
\begin{equation}\label{eq:loss_d_f}
    \begin{aligned}
        \mathcal{L}_{adv}=\frac{1}{B}\sum_{i=1}^B\left(\frac{1}{T}\sum_{t=1}^T\mathcal{L}_{adv}^{i,t}\right)=\frac{1}{B\cdot T}\sum_{i=1}^B\sum_{t=1}^T\mathcal{L}_{adv}^{i,t},
    \end{aligned}
\end{equation}
where $B$ is the batch size. 

\subsection{Stochastic Step Distillation}
As shown in Fig.~\ref{fig:framwork}, in addition to an adversarial loss $\mathcal{L}_{adv}$, there is also distillation loss between the outputs of the student model and the teacher model, as well as between the outputs of the student model and the original data. For this process, we adhere to the design and training method delineated in~\cite{hinton2015distilling}, a summary of which is provided herein. In time step $t$, we can formulate the loss as follows:
\begin{equation}
    \begin{aligned}
        \mathcal{L}_{dis}^{i,t}=\mathcal{L}_{MSE}(x_{\psi,i,t-1}, x_{\theta,i,t-1})+\mathcal{L}_{MSE}(x_{i,t-1}, x_{\theta,i,t-1}),
    \end{aligned}
\end{equation}
where $\mathcal{L}_{MSE}$ represents the mean squared error~(MSE) loss. The distillation loss~($\mathbb{E}_d(\epsilon_\theta;\epsilon_\psi,\epsilon_\phi)$) for a batch of data over the entire time steps $T$ is given as:
\begin{equation}\label{eq:loss_s}
    \begin{aligned}
        \mathcal{L}_{dis}=\frac{1}{B}\sum_{i=1}^B\left(\frac{1}{T}\sum_{t=1}^T\mathcal{L}_{dis}^{i,t}\right)=\frac{1}{B\cdot T}\sum_{i=1}^B\sum_{t=1}^T\mathcal{L}_{dis}^{i,t}.
    \end{aligned}
\end{equation}

Based on the above analysis, we incorporate the adversarial loss $\mathcal{L}_{adv}$ in Eq.~(\ref{eq:loss_d_f}) and the distillation loss $\mathcal{L}_{dis}$ in Eq.~(\ref{eq:loss_s}) into our final loss function. Our whole framework is trained end-to-end by the following objective function:
\begin{equation}
    \begin{aligned}
        \mathcal{L}=\mathcal{L}_{dis} + \lambda \mathcal{L}_{adv},
    \end{aligned}
\end{equation}
where $\lambda$ is a trade-off weight. We set it as $1$ in our experiments. 

Without any protection, we calculate the gradients for backpropagation as follows:
\begin{equation}
    \begin{aligned}
        g = \frac{\partial \mathcal{L}}{\partial \theta}
          = \frac{1}{B\cdot T}\sum_{i=1}^B\sum_{t=1}^T\left(\frac{\partial(\mathcal{L}_{dis}^{i,t}+\lambda\mathcal{L}_{adv}^{i,t})}{\partial\theta}\right).
    \end{aligned}
\end{equation}
Directly updating the student model with gradients $g$ may lead to privacy leakage. Therefore, we implement differential privacy protection by clipping it and adding noise. The specific process is as follows:
\begin{equation}\label{eq:dp_g}
    \begin{aligned}
        \bar{g} &= \frac{1}{B\cdot T}\left(\sum_{i=1}^B\sum_{t=1}^T CLIP\left(\frac{\partial(\mathcal{L}_{dis}^{i,t}+\lambda\mathcal{L}_{adv}^{i,t})}{\partial\theta},C\right)+\mathcal{N}(0,\sigma^2C^2\textbf{I})\right)\\
        &= \frac{1}{B\cdot T}\sum_{i=1}^B\sum_{t=1}^T\left(CLIP\left(\frac{\partial(\mathcal{L}_{dis}^{i,t}+\lambda\mathcal{L}_{adv}^{i,t})}{\partial\theta},C\right)\right)+\frac{\mathcal{N}(0,\sigma^2C^2\textbf{I})}{B\cdot T}\\
        &= \frac{1}{B}\sum_{i=1}^B\left(\frac{1}{T}\sum_{t=1}^T\left(CLIP\left(\frac{\partial(\mathcal{L}_{dis}^{i,t}+\lambda\mathcal{L}_{adv}^{i,t})}{\partial\theta},C\right)\right)\right)+\frac{\mathcal{N}(0,\sigma^2C^2\textbf{I})}{B\cdot T},
    \end{aligned}
\end{equation}
where $CLIP(*,C)=*/max(1,\frac{||*||_2}{C})$. In our experiments, it was observed that an increase in the value of $T$ correlates with an enhancement in data quality. Nonetheless, as dictated by Eq~(\ref{eq:dp_g}), each sample is subjected to $T$ steps of diffusion throughout the training process, resulting in inefficiencies. To mitigate this, we substitute the average of the gradients over all $T$ time steps with the gradient from a randomly selected time step.
\begin{equation}\label{eq:dp_r}
    \begin{aligned}
    \bar{g}\approx\frac{1}{B}\sum_{i=1}^B\left(CLIP\left(\frac{\partial(\mathcal{L}_{dis}^{i,r}+\lambda\mathcal{L}_{adv}^{i,r})}{\partial\theta},C\right)\right)+\frac{\mathcal{N}(0,\sigma^2C^2\textbf{I})}{B\cdot T},
    \end{aligned}
\end{equation}
where $r$ is a number randomly selected from 0 to $T$. Compared to existing methods~\cite{Dockhorn2022DifferentiallyPD,ghalebikesabi2023differentially,Saiyue2023dpldm} that directly employ DPSGD to train diffusion models, our method uses the time step $T$ to dilute the impact of noise without compromising privacy protection. 

Based on previous work~\cite{chen2020gs}, we find that directly clipping and adding noise to each gradient introduces more randomness, leading to a decrease in the convergence speed of training. According to the properties of the chain rule for gradients, we have:
\begin{equation}
    \begin{aligned}
    \frac{\partial(\mathcal{L}_{dis}^{i,r}+\lambda\mathcal{L}_{adv}^{i,r})}{\partial\theta}=\frac{\partial(\mathcal{L}_{dis}^{i,r}+\lambda\mathcal{L}_{adv}^{i,r})}{\partial x_{\theta,i,r}}\cdot \frac{\partial x_{\theta,i,r}}{\partial \theta}.
    \end{aligned}
\end{equation}
Combining the post-processing property of differential privacy, we can modify Eq~(\ref{eq:dp_r}) as follows:
\begin{equation}\label{eq:dp_f}
    \begin{aligned}
    \bar{g}\approx\frac{1}{B}\sum_{i=1}^B\left(CLIP\left(\frac{\partial(\mathcal{L}_{dis}^{i,r}+\lambda\mathcal{L}_{adv}^{i,r})}{\partial x_{\theta,i,r-1}},C\right)\cdot\frac{\partial x_{\theta,i,r-1}}{\partial \theta}\right)+\frac{\mathcal{N}(0,\sigma^2C^2\textbf{I})}{B\cdot T}.
    \end{aligned}
\end{equation}
By truncating randomness in this manner, we only need to introduce randomness to $x_{\theta,i,r-1}$ once to achieve the same level of privacy protection.

\subsection{Privacy Analysis}
In this section, we analyze the differential privacy bound for our proposed DP-SAD and we leverage the Renyi differential privacy~(RDP)~\cite{mironov2017renyi} and Gaussian mechanism~\cite{dwork2014algorithmic} in our analysis.

\begin{definition}[R$\acute{\textbf{e}}$nyi Differential Privacy]\label{def:rdp}
\itshape{A randomized mechanism $\mathcal{A}$ is $(q, \varepsilon)$-RDP with $q > 1$ if for any adjacent datasets $\mathcal{D}$ and $\mathcal{D}'$ :}
\begin{equation}
\begin{aligned}
D_{q}&(\mathcal{A}(\mathcal{D})||\mathcal{A}(\mathcal{D}'))=\frac{1}{q-1}\log \mathbb{E}_{(x\sim\mathcal{A}(\mathcal{D}))}\left[\left(\frac{\operatorname{Pr}[\mathcal{A}(\mathcal{D})=x]}{\operatorname{Pr}[\mathcal{A}(\mathcal{D}')=x]}\right)^{q-1}\right]\le \varepsilon.
\end{aligned}
\end{equation}
\end{definition}
\begin{theorem}[Convert RDP to DP]\label{th:rdp-dp}
\itshape{A $(q, \varepsilon)$-RDP mechanism $\mathcal{A}$ also satisfies $(\varepsilon+\log \frac{q-1}{q}-\frac{\log \delta + \log q}{q-1}, \delta)$-DP.}
\end{theorem}
\begin{theorem}[Gaussian Mechanism]\label{th:gaussian}
\itshape{Let $f$ be a function with sensitive being $S_f=\max\limits_{\mathcal{D},\mathcal{D}^{\prime}}||f(\mathcal{D})-f(\mathcal{D}^{\prime})||_2$ over all adjacent datasets $\mathcal{D}$ and $\mathcal{D}^{\prime}$. The Gaussian mechanism $\mathcal{A}$ with adding noise to the output of $f$:$\mathcal{A}(x) = f(x) + \mathcal{N}(0, \sigma^2)$
is $(q, \frac{q S_f^{2}}{2\sigma^2})$-RDP.}
\end{theorem}

We first calculate the sensitivity of the function that implements differential privacy. Then, based on the definitions and theories mentioned above, we derive the privacy bound of our DP-SAD.
\begin{lemma}\label{lemma:sensitive}
\itshape{For any neighboring gradient vectors $\bar{g}, \bar{g}^{\prime}$ differing by the gradient vector of one data with length $s$, the $l_2$ sensitivity is $2C\sqrt{s}$ after performing normalization with normalization bound $C$.}
\begin{proof}
\itshape{The $l_2$ sensitivity is the max change in $l_2$ norm caused by the input change. For the vectors after normalization with norm bound $C$, each dimension has a maximum value of $C$ and a minimum value of $-C$. In the worst case, the difference of one data makes the gradient of all dimensions change from the maximum value $C$ to the minimum value $-C$, the change in $l_2$ norm equals $\sqrt{(2C)^2s}=2C\sqrt{s}$.}
\end{proof}
\end{lemma}
We assume that the batch size is $B$, the number of iterations is $N$, and the variance of the noise added each time is $\sigma^2$.
\begin{theorem}\label{th:dp-proof}
\itshape{DP-SAD guatantees $(\frac{2C^2sBN\lambda}{\sigma^2}+\log \frac{\lambda-1}{\lambda}-\frac{\log \delta + \log \lambda}{\lambda-1},\delta)$-DP for all $\lambda\ge 1$ and $\delta\in(0,1)$.}
\begin{proof}
For each data, the gradient clipping and noise addition implements a Gaussian mechanism which guarantees $(\lambda, \frac{2C^2s\lambda}{\sigma^2})$-RDP (Theorem~\ref{th:gaussian} \& Lemma~\ref{lemma:sensitive}). So the DP-SAD satisfies $(\lambda, \frac{2C^2sBN\lambda}{\sigma^2})$-RDP, which is $(\frac{2C^2sBN\lambda}{\sigma^2}+\log \frac{\lambda-1}{\lambda}-\frac{\log \delta + \log \lambda}{\lambda-1},\delta)$-DP~(Theorem~\ref{th:rdp-dp}).
\end{proof}
\end{theorem}

\section{Experiments}

To verify the effectiveness of our proposed DP-SAD, we compare it with 11 state-of-the-art methods and evaluate the data utility and visual quality on three image datasets. To ensure fair comparisons, our experiments adopt the same settings as these baselines and cite results from their original papers.

\subsection{Experimental Setup}
In this section, we provide a brief description of the experimental settings. For more in-depth experimental details, please refer to the supplementary material.

\myPara{Datasets.}~We conduct experiments on three image datasets, including MNIST~\cite{Lecun98cnn}, FashionMNIST~(FMNIST)~\cite{fashionMnist} and CelebA~\cite{celeba}. To further refine our analysis, we derive two subsets from CelebA, namely CelebA-H and CelebA-G, which are created with hair color~(black/blonde/brown) and gender as the label.

\myPara{Baselines.}~We compare our DP-SAD with 11 state-of-the-art methods, including DP-GAN~\cite{xie2018differentially}, PATE-GAN~\cite{jordon2018pate}, DP-MERF~\cite{harder2020differentially}, GS-WGAN~\cite{chen2020gs}, P3GM~\cite{takagi2021p3gm}, G-PATE~\cite{long2021g}, DataLens~\cite{wang2021ccs}, DPGEN~\cite{chen2022dpgen}, PSG~\cite{chen2022privateset}, DP-DM~\cite{Dockhorn2022DifferentiallyPD} and DP-LDM~\cite{Saiyue2023dpldm}. 

\myPara{Metrics.}~We evaluate our DP-SAD as well as baselines in terms of perceptual scores and classification accuracy under the same different privacy budget constraints. In particular, perceptual scores are evaluated by Inception Score~(IS) and Frechet Inception Distance~(FID), which are standard metrics for the visual quality of images. Classification accuracy is evaluated by training a classifier with the generated data and testing it on real test datasets

\begin{figure}[!t]
	\centering
	\includegraphics[width=0.9\linewidth]{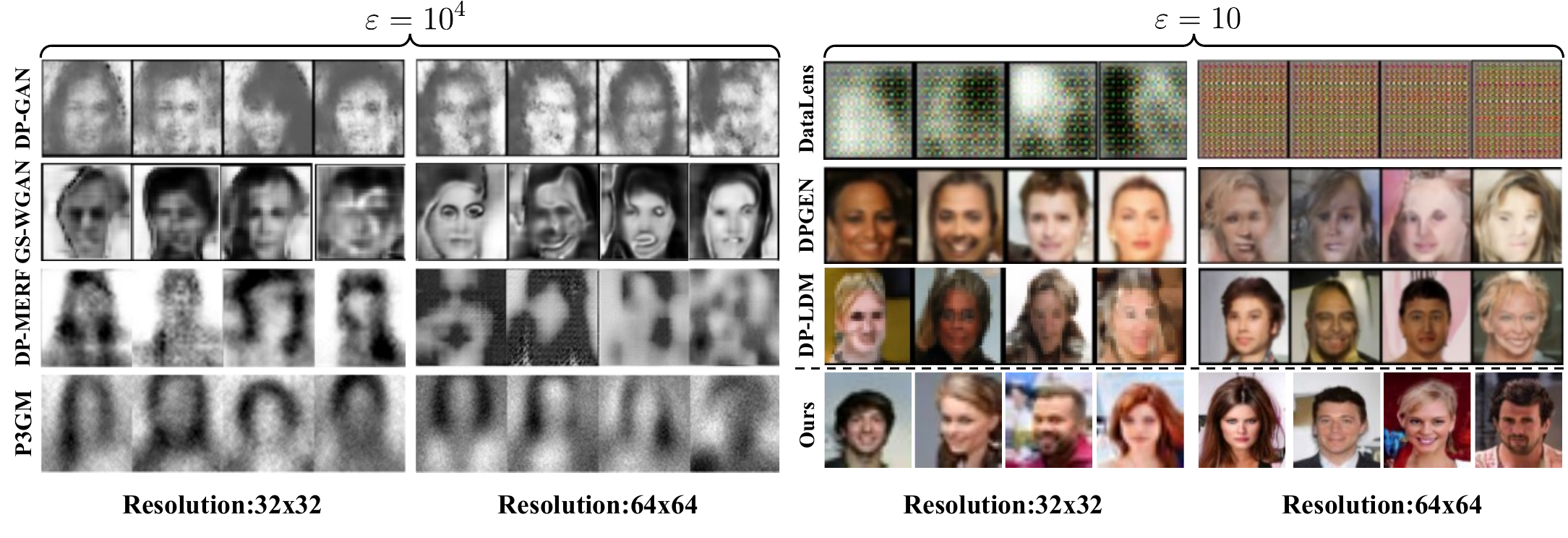}
	\caption{Visualization results of DP-GAN, GS-WGAN, DP-MERF, P3GM, DataLens, DPGEN, DP-LDM and our DP-SAD on CelebA at 32$\times$32 and 64$\times$64 resolutions.}
	\label{fig:visual-compare}
\end{figure}

\begin{figure}[t]
	\centering
	\includegraphics[width=1\linewidth]{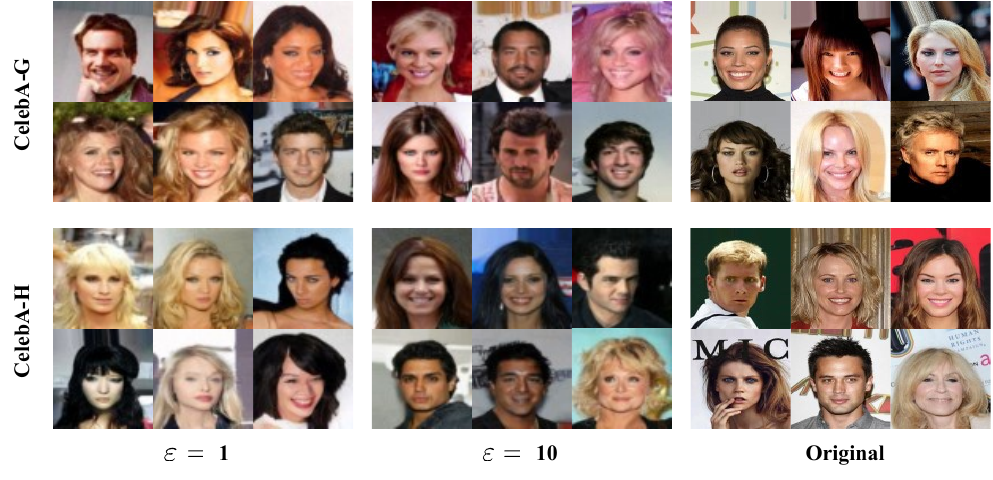}
	\caption{Generated samples by DP-SAD on CelebA-G and CelebA-H under different privacy budget~($\varepsilon=1$ and $\varepsilon=10$).}
	\label{fig:celeba}
\end{figure}

\begin{table}[t]
\small
 \setlength{\tabcolsep}{10.5pt}%
 \renewcommand\arraystretch{1.0}
    \caption{Perceptual scores comparisons with 9 state-of-the-art baselines on CelebA at 64 $\times$ 64 resolution under different privacy budget $\varepsilon$.}\label{tab:score}
	\begin{center}
		\begin{threeparttable}
			\begin{tabular}{l|lcc}
				\textbf{Method}& ~~$\varepsilon$   &   \textbf{IS}$\uparrow$  &   \textbf{FID}$\downarrow$\cr
				\hline
                \textit{Without pre-training} &&&\cr
				\textbf{DP-GAN} (arXiv'18)  &   $10^4$    &   1.00    &   403.94\cr
				\textbf{PATE-GAN} (ICLR'19)&   $10^4$    &   1.00    &   397.62\cr
				\textbf{GS-WGAN} (NeurIPS'20) &   $10^4$    &   1.00    &   384.78\cr
				\textbf{DP-MERF} (AISTATS'21) &   $10^4$    &   1.36    &   327.24\cr
				\textbf{P3GM} (ICDE'21)    &   $10^4$    &   1.37    &   435.60\cr
				\textbf{G-PATE} (NeurIPS'21)  &   $10$      &   1.37    &   305.92\cr
				\textbf{DataLens} (CCS'21)&   $10$      &   1.42    &   320.84\cr
				\textbf{DPGEN} (CVPR'22)  &   $10$      &   1.48    &   55.910\cr
                \hline
                \textit{With pre-training} &&&\cr
                \textbf{DP-LDM} (arXiv'23)&   $10$      &   N/A    &   14.300\cr
                \hline
				\textbf{DP-SAD} (Ours)&$10$    &   \textbf{2.37}    &   \textbf{11.260}\cr
			\end{tabular}
		\end{threeparttable}
	\end{center}
\end{table}

\begin{table}[t]
\small
 \setlength{\tabcolsep}{1.5pt}%
 \renewcommand\arraystretch{1.0}
   \caption{Classification accuracy comparisons with 11 state-of-the-art baselines under different privacy budget $\varepsilon$.}\label{tab:dp(1_and_10)}
	\begin{center}
		\begin{threeparttable}
			\begin{tabular}{l|cccccccccccc}
				\multirow{2}{*}{\textbf{Method}}&&\multicolumn{2}{c}{\textbf{MNIST}} &&\multicolumn{2}{c}{\textbf{FMNIST}}&&\multicolumn{2}{c}{\textbf{CelebA-H}}&&\multicolumn{2}{c}{\textbf{CelebA-G}}\cr
                 && $\varepsilon$=1 & $\varepsilon$=10 && $\varepsilon$=1 & $\varepsilon$=10 && $\varepsilon$=1 & $\varepsilon$=10 && $\varepsilon$=1 & $\varepsilon$=10\cr
                \hline
                \textit{Without pre-training} &&&&&&&&&&&&\cr
                \textbf{DP-GAN} && 0.4036 & 0.8011 && 0.1053 & 0.6098 && 0.5330 & 0.5211 && 0.3447 & 0.3920\cr
                \textbf{PATE-GAN} && 0.4168 & 0.6667 && 0.4222 & 0.6218 && 0.6068 & 0.6535 && 0.3789 & 0.3900\cr
                \textbf{GS-WGAN} && 0.1432 & 0.8075 && 0.1661 & 0.6579 && 0.5901 & 0.6136 && 0.4203 & 0.5225\cr
                \textbf{DP-MERF} && 0.6367 & 0.6738 && 0.5862 & 0.6162 && 0.5936 & 0.6082 && 0.4413 & 0.4489\cr
                \textbf{P3GM} && 0.7369 & 0.7981 && 0.7223 & 0.7480 && 0.5673 & 0.5884 && 0.4532 & 0.4858\cr
                \textbf{G-PATE} && 0.5810 & 0.8092 && 0.5567 & 0.6934 && 0.6702 & 0.6897 && 0.4985 & 0.6217\cr
                \textbf{DataLens} && 0.7123 & 0.8066 && 0.6478 & 0.7061 && 0.7058 & 0.7287 && 0.6061 & 0.6224\cr
                \textbf{DPGEN} && 0.9046 & 0.9357 && 0.8283 & 0.8784 && 0.6999 & 0.8835 && 0.6614 & 0.8147\cr
                \textbf{PSG} && 0.8090 & 0.9560 && 0.7020 & 0.7770 && N/A & N/A && N/A & N/A\cr
                \hline
                \textit{With pre-training} &&&&&&&&&&&&\cr
                \textbf{DP-DM} && 0.9520 & \textbf{0.9810} && 0.7940 & 0.8620 && N/A & N/A && N/A & N/A\cr
                \textbf{DP-LDM} && 0.9590 & 0.9740 && N/A & N/A && N/A & N/A && N/A & N/A\cr
                \hline
                \textbf{DP-SAD} (Ours) && \textbf{0.9621} & 0.9761 && \textbf{0.8437} & \textbf{0.8960} && \textbf{0.9150} & \textbf{0.9280} && \textbf{0.8263} & \textbf{0.8414}\cr
			\end{tabular}
		\end{threeparttable}
	\end{center}
\end{table}

\subsection{Experimental Results}

\myPara{Visual comparisons of generated data.}~We furnish visual evidence to substantiate the superior quality of data generated through our method. In Fig.~\ref{fig:visual-compare}, we juxtapose our visualization outcomes against those derived from other benchmark models. Notably, even when operating under a stringent privacy budget ($\varepsilon = 10^4$), the grayscale images produced by DP-GAN, GS-WGAN, DP-MERF, and P3GM exhibit a noticeable degree of blurriness. We underscore the intrinsic advantage of grayscale images, which, due to their reduced dimensionality, facilitate a more manageable equilibrium between data quality and privacy preservation. In contrast, the color images generated by DPGEN and DP-LDM showcase a higher visual quality relative to DataLens, albeit with a lack of detailed facial features. Against this backdrop, the images emanated from our DP-SAD distinguish themselves by presenting a more lifelike appearance coupled with enhanced facial detail, thereby validating the efficacy of our DP-SAD.

\myPara{Image generated by DP-SAD.}~We present the visual quality evaluation results in Fig.~\ref{fig:celeba}, where all of the images were generated by DP-SAD. We find that samples at $\varepsilon=10$ possess more facial details compared to samples at $\varepsilon=1$. Compared to the images of 64$\times$64 resolution presented in Fig.~\ref{fig:visual-compare}, the results of DP-SAD on images of the same resolution manifest a significantly enhanced realism and display a markedly improved facial structure. This observation highlights the ability of DP-SAD to produce more lifelike and structurally accurate facial images even under more stringent privacy settings, further evidencing the superior performance of our proposed method in generating high-quality images.

\myPara{Perceptual scores comparisons.}~To further substantiate the efficacy of our DP-SAD, we conducted evaluations using two established metrics: IS and FID, as previously discussed. Due to the absence of experimental data for PSG and DP-DM, our comparative analysis was limited to the remaining 9 methods. The outcomes of this comparison are detailed in Tab.~\ref{tab:score}. A superior IS value is indicative of enhanced quality in the generated samples, whereas a diminished FID score suggests a closer resemblance to authentic images. Among the evaluated baselines, our technique distinguished itself by recording the highest IS value of 2.37 and the lowest FID score of 11.260 under the most restrictive privacy budget of 10. This can be attributed to two factors: one is that we utilized the diffusion time steps to dilute the impact of DP noise, and the other is that we incorporated a discriminator to form adversarial training. The combination of these two aspects has improved the performance of the model.

\myPara{Downstream task performance comparison.}~Furthermore, we compare our DP-SAD with existing DP generative methods on classification tasks under two privacy budget settings $\varepsilon=1$ and $\varepsilon=10$ on MNIST, FMNIST, CelebA-H and CelebA-G. We evaluate the classification accuracy of the classifiers trained on the generated data, and the results are summarized in Tab.~\ref{tab:dp(1_and_10)}. It is important to note that our method does not require pre-training. 
Compared to methods without pre-training, we observe consistent and significant improvements of around 4-6 percentage points across different configurations. Especially for complex tasks~(CelebA) where $\varepsilon=1$, our method outperforms other methods by at least 16 percentage points. Furthermore, compared to the two methods with pre-training, our method consistently achieves optimal results in most settings. This improvement is attributed to the fact that we chose a more stable diffusion model instead of GANs and diluted the impact of DP noise with time steps to achieve a better balance between privacy and utility. These results suggest that DP-SAD can effectively generate high-quality images with practical applications.

    

\begin{figure}[!t]
	\centering
	\includegraphics[width=0.49\linewidth]{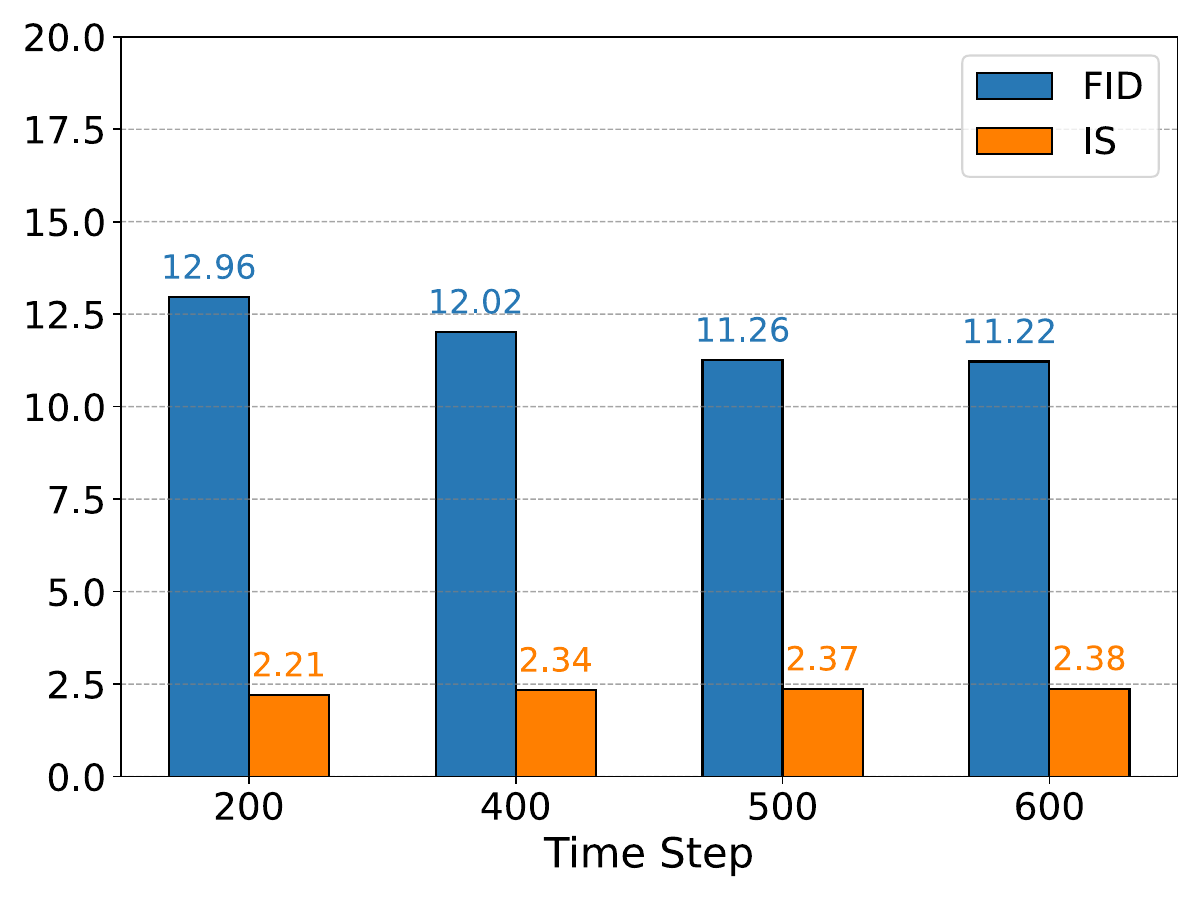}
        \includegraphics[width=0.49\linewidth]{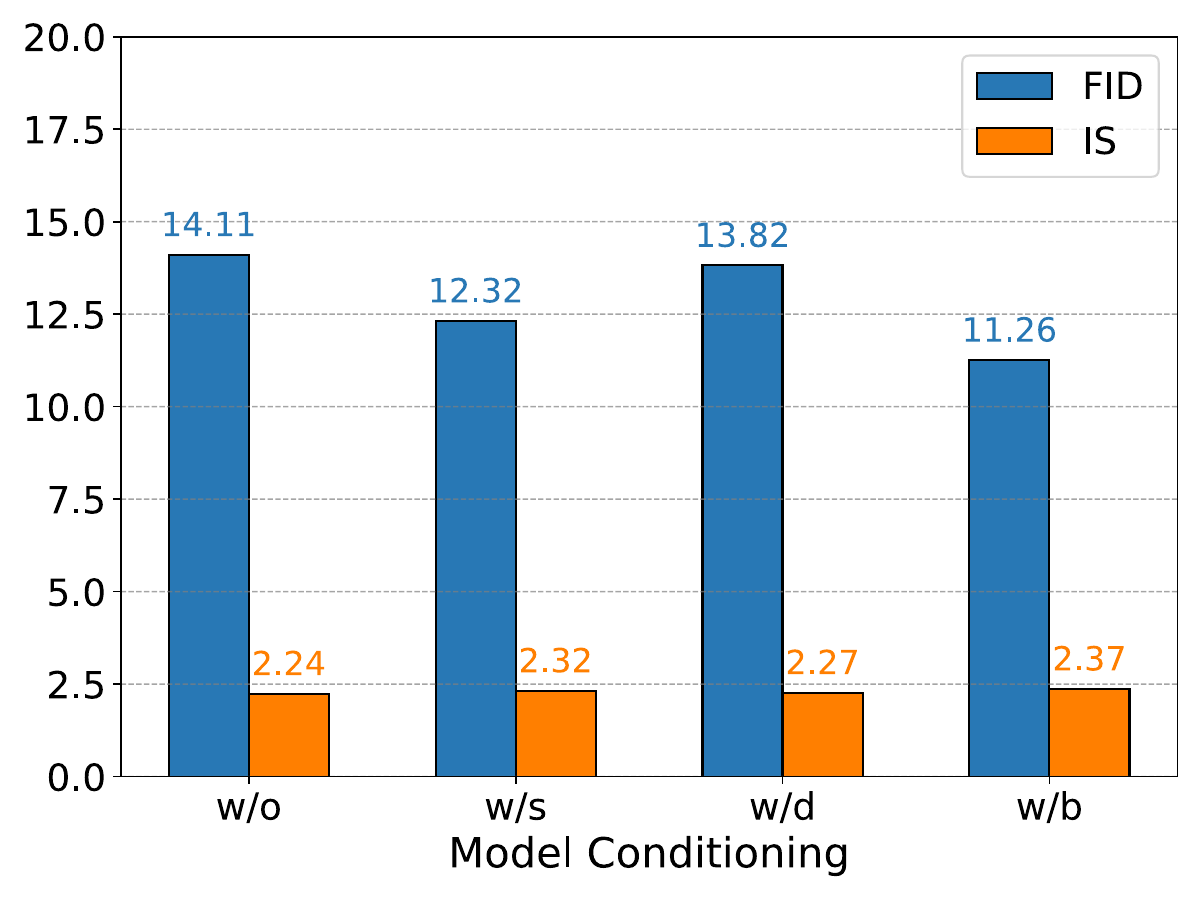}
	\caption{Left: Perceptual scores on CelebA under different time steps. Right: Perceptual scores on CelebA under different model conditioning settings~(w/o: without model conditioning, w/s: with student conditioning, w/d: with discriminator conditioning, w/b: with both model conditioning).}
	\label{fig:ab}
\end{figure}


\begin{table}[t]
\setlength{\tabcolsep}{6.0pt}%
\renewcommand\arraystretch{1.0}
\caption{Perceptual scores on CelebA under different trade-off weight $\lambda$.}\label{tab:lam}
\begin{center}
\begin{threeparttable}
\begin{tabular}{l|cccccc}
    $\lambda$ & 0.0 & 0.2 & 0.5 & 1.0 & 2.0 & 4.0\cr
    \hline
    \textbf{FID} & 14.63 & 13.41 & 12.38 & 11.68 & 12.11 & 13.55\cr
    \hline
    \textbf{IS} & 2.12 & 2.18 & 2.26 & 2.37 & 2.35 & 2.31 \cr
\end{tabular}\end{threeparttable}\end{center}
\end{table}

\subsection{Ablation Studies}
After the promising performance is achieved, we further analyze the impact of each component of our method, including the time step $T$, the model conditioning and the trade-off weight $\lambda$.

\myPara{Impact of time step.}~To investigate the impact of the time step on the trade-off between privacy and data utility, we compare the perceptual scores obtained when the time step $T$ takes different values under the same privacy budget $\varepsilon=10$. The results are shown in Fig.~\ref{fig:ab} left. As we anticipated, with the increase in $T$, the IS increases and the FID decreases, which collectively indicates an improvement in image quality. This is because, on one hand, as demonstrated by Eq.~(\ref{eq:dp_f}), an increase in $T$ will reduce the influence of noise on the gradient; on the other hand, an increase in $T$ will enhance the generative effect of the diffusion model itself. Although increasing $T$ improves performance, it reduces training efficiency. Therefore, in other parts of experiments, we choose $T=500$.

\myPara{Impact of model conditioning.}~To explore the effect of the model conditioning, including student conditioning and discriminator conditioning, we conduct experiments with/without model conditioning under the same privacy budget $\varepsilon=10$. The results are shown in Fig.~\ref{fig:ab} right. We observe that model conditioning enhances results. Notably, student conditioning outperforms discriminator conditioning, and the combination of both student conditioning and discriminator conditioning yields the best results. An additional benefit of student conditioning is that the data inherently comes with labels when conducting downstream tasks, e.g., classifier training. Labeling data through a pre-trained model~(trained with private data without any protection) may lead to privacy leakage.

\myPara{Impact of $\lambda$.}~To study the effect of $\lambda$  on the quality of the generated images, we train the student model with different trade-off weight $\lambda$ under the same privacy budget $\varepsilon=10$. The results are presented in Tab.~\ref{tab:lam}. As $\lambda$ increases from 0 to 1, the image quality improves with the increase in $\lambda$, as the discriminative loss $\mathcal{L}_{adv}$ drives the output distribution of the student model closer to that of the teacher model. However, when $\lambda$ exceeds 1, the image quality decreases with the increase in $\lambda$. We speculate that this may be due to the larger discriminative loss $\mathcal{L}_{adv}$ constraining the efficacy of $\mathcal{L}_{dis}$. This also inspires our subsequent work, suggesting that assigning different values to $\lambda$ at different time steps might be more beneficial for model training.

\section{Conclusion}
Direct data sharing may pose the risk of privacy leakage. To address this challenge, we proposed DP-SAD, a differentially private generative model trained by a stochastic adversarial distillation method. It achieves differential privacy by clipping the gradients and adding noise. We ingeniously dilute the impact of noise through the diffusion model's time steps and incorporate a discriminator to form adversarial training with the student model. This method endows our model with superior performance compared to other methods. Furthermore, we combine the chain rule of gradients with the post-processing property of differential privacy to reduce the introduction of randomness, which accelerates the entire training process. Extensive experiments and analysis clearly demonstrate the effectiveness of our proposed method.

\section*{Acknowledgements}
This work was partially supported by grants from the Pioneer R\&D Program of Zhejiang Province (2024C01024), and Open Research Project of the State Key Laboratory of Media Convergence and Communication, Communication University of China (SKLMCC2022KF004).

%
%
\bibliographystyle{splncs04}
\bibliography{egbib}

\section{Appendix}
\subsection*{Procedure of DP-SAD}
The procedure of our DP-SAD is shown in Alg.~\ref{alg:DP-SAD}.  It can be described as the following four steps:
\begin{itemize}
    \item train a teacher model $\epsilon_\psi$ without protection.
    \item initialize the student model $\epsilon_\theta$ and the discriminator $\epsilon_\phi$.
    \item randomly sample a batch of data :~$\left\{x_i\right\}^{B}_{i=1}$.
    \item calculate the loss function with Eq.~(12) and update the student and the discriminator with Eq.~(17).
\end{itemize}
We run the last two steps until the termination condition is reached.

\begin{algorithm}[!htbp]
\small
\caption{\small{DP-SAD}}
\label{alg:DP-SAD}
\begin{algorithmic}[1] 
\REQUIRE Private data $\mathcal{D}$, time step $T$, batch size $B$, training iterations $N$, learning rates $\gamma$ and $\gamma_d$, the teacher $\epsilon_\psi$, the student $\epsilon_\theta$ and the discriminator $\epsilon_\phi$.
\STATE Train a teacher model with private data $\mathcal{D}$ without protection.
\STATE Initialize $\theta_0$ and $\phi_0$ with Xavier.
\FOR {$k$ < $N$}
\STATE sample a batch of data from $\mathcal{D}$ :~$\left\{x_i\right\}^{B}_{i=1}$.
\STATE Compute the final loss $\mathcal{L}$ with Eq.~(12) and get the differentially private gradients $\bar{g}$ with Eq.~(17).
\STATE Update the student with $\theta_{k+1}=\theta_k-\gamma\cdot\bar{g}$
\STATE Compute the loss $\mathcal{L}_r=\frac{1}{B}\sum_{i=1}^B\mathcal{L}_{adv}^{i,r}$ and update the discriminator with $\phi_{k+1}=\phi_k-\gamma_d\cdot\partial \mathcal{L}_r/\partial\phi$.
\ENDFOR
\RETURN $\theta_N$
\end{algorithmic}
\end{algorithm}

\subsection*{Convergence Analysis}
We assume that the function to be optimized is $\mathcal{L}(\theta)$, where $\theta$ is the parameter of the student model. We follow the standard assumptions same as~\cite{bottou2018optimization}:
\begin{equation}
    \begin{aligned}
        &(1)~||\nabla \mathcal{L}(\theta)-\nabla \mathcal{L}(\theta^{\prime})||_2\leq \tau_1||\theta-\theta^{\prime}||_2;\\
        &(2)~\mathcal{L}(\theta) \geq \mathcal{L}(\theta^{\prime}) + \nabla \mathcal{L}(\theta^{\prime})^{T}(\theta-\theta^{\prime})+\frac{\tau_2}{2}||\theta-\theta^{\prime}||_2^2;\\
        &(3)~\nabla \mathcal{L}(\theta)^T\mathbb{E}_t[\bar{g}(\theta;t)]\geq \mu||\nabla \mathcal{L}(\theta)||_2^2,
    \end{aligned}
\end{equation}
where $\nabla\mathcal{L}(\theta)$ is the true gradient~(Eq~(13) in the main text), $\bar{g}(\theta;t)$ is the gradient we used to update the student~(Eq~(15) in the main text), $\mathbb{E}[\cdot]$ is the symbol for mean calculation, $\mathbb{V}[\cdot]$ is the symbol for variance calculation and $\tau_1,\tau_2,\mu,\mu_e,\mu_v,c$ are non-negative constants. According to assumption (1), we have:
\begin{lemma}\label{lemma:lipschitz}
    \itshape{For any two weights $\theta$ and $\theta^{\prime}$, the difference of the objective function $\mathcal{L}(\theta)-\mathcal{L}(\theta^{\prime})$ is limited by the distance between the weights.
    \begin{equation}
        \begin{aligned}
            \mathcal{L}(\theta)\leq \mathcal{L}(\theta^{\prime})+\nabla \mathcal{L}(\theta^{\prime})^T(\theta-\theta^{\prime})+\frac{\tau_1}{2} ||\theta-\theta^{\prime}||_2^2.
        \end{aligned}
    \end{equation}
    }
    \begin{proof}
        The Taylor expansion of the objective function $\mathcal{L}(\theta)$ can be expressed as:
        \begin{equation}\label{eq:taylor}
            \begin{aligned}
                \mathcal{L}(\theta) = \mathcal{L}(\theta^{\prime}) + \nabla \mathcal{L}(\theta^{\prime})^{T}(\theta-\theta^{\prime})+\frac{1}{2}(\theta-\theta^{\prime})^T\nabla^2\mathcal{L}(\vartheta)(\theta-\theta^{\prime}),
            \end{aligned}
        \end{equation}
        where $\vartheta$ is any point between $\theta$ and $\theta^{\prime}$. According to assumption (1), we know the Hessian matrix satisfies:
        \begin{equation}\label{eq:hessian}
            \begin{aligned}
                \nabla^2\mathcal{L}(\theta) \leq \tau_1.
            \end{aligned}
        \end{equation}
        Combining Eq.~(\ref{eq:taylor}) and Eq.~(\ref{eq:hessian}), we get Lemma.~\ref{lemma:lipschitz}.
    \end{proof}
\end{lemma}

Based on the assumption (2), we have:
\begin{lemma}\label{lemma:as2}
    \itshape{For any weight $\theta$, the distance between $\mathcal{L}(\theta)$ and the minimum value $\mathcal{L}(\theta^{*})$ is limited by $\nabla \mathcal{L}(\theta)$ as follows
    \begin{equation}
        \begin{aligned}
            \mathcal{L}(\theta)-\mathcal{L}(\theta^{*})\leq \frac{1}{2\tau_2}||\nabla \mathcal{L}(\theta)||_2^2.
        \end{aligned}
    \end{equation}
    }
    \begin{proof}
        We regard the right side of the inequality as a quadratic function on $\theta$. When $\theta=\theta^{\prime}-\frac{1}{\tau_2}\nabla \mathcal{L}(\theta^{\prime})$, it takes the minimum value $\mathcal{L}(\theta^{\prime})-\frac{1}{2\tau_2}||\nabla \mathcal{L}(\theta^{\prime})||_2^2$. Substituting it into assumption (2) and letting $\theta=\theta^{*}$, we can get Lemma~\ref{lemma:as2}.
    \end{proof}
\end{lemma}

We consider the update at step $k$ as $\theta_{k+1}=\theta_k - \gamma\cdot \bar{g}(\theta_k;t)$. Based on Lemma.~\ref{lemma:lipschitz}, we have:
\begin{equation}
    \begin{aligned}
        \mathcal{L}(\theta_{k+1})\leq \mathcal{L}(\theta_k)-&\gamma\nabla \mathcal{L}(\theta_k)^T\bar{g}(\theta_k;t)
            +\frac{\tau_1}{2}\gamma^2||g(\theta_k;t)||_2^2.
    \end{aligned}
\end{equation}
Taking the expectations on both sides gives:
\begin{equation}\label{eq:expect}
    \begin{aligned}
        \mathbb{E}[\mathcal{L}(\theta_{k+1})-\mathcal{L}(\theta_k)]\leq-\gamma\nabla \mathcal{L}(\theta_k)^T\mathbb{E}[\bar{g}(\theta_k,t)]+\frac{\tau_1}{2}\gamma^2\mathbb{E}[||\bar{g}(\theta_k;t)||^2_2].
    \end{aligned}
\end{equation}
Since $\bar{g}=CLIP(g,C)=g/max(1,\frac{||g||_2}{C})+\mathcal{N}(0,\sigma^2C^2\textbf{I})$, combining with Cauchy Schwartz inequality yields:
\begin{equation}\label{eq:Cauthy}
    \begin{aligned}
        \mathbb{E}[||\bar{g}(\theta; t)||_2^2] \leq 2C^2 + 2\sigma^2C^2d,
    \end{aligned}
\end{equation}
where $d=||z||^2$, $z\sim\mathcal{N}(0,\textbf{I})$. Substituting Eq.~(\ref{eq:Cauthy}) into Eq.~{\ref{eq:expect}} and combining it with assumption (3) yields:
\begin{equation}
    \begin{aligned}
        \mathbb{E}[\mathcal{L}(\theta_{k+1})-\mathcal{L}(\theta_k)]\leq-\gamma\mu||\nabla \mathcal{L}(\theta)||_2^2+\frac{\tau_1}{2}\gamma^2(2C^2 + 2\sigma^2C^2d).
    \end{aligned}
\end{equation}
Combined with Lemma.~\ref{lemma:as2}, we have:
\begin{equation}
    \begin{aligned}
        \mathbb{E}[\mathcal{L}(\theta_{k+1})-\mathcal{L}(\theta_k)]\leq-2\tau_2\gamma\mu\mathbb{E}[\mathcal{L}(\theta_k)-\mathcal{L}(\theta^{*})]+\frac{\tau_1}{2}\gamma^2(2C^2 + 2\sigma^2C^2d).
    \end{aligned}
\end{equation}
We define $d_k = \mathcal{L}(\theta_k)-\mathcal{L}(\theta^{*})$ and after the transformation we have:
\begin{equation}
    \begin{aligned}
        d_{k+1} - \frac{\tau_1\gamma C^2(1+\sigma^2d)}{2\tau_2\mu}\leq(1-2\tau_2\gamma\mu)(d_k-\frac{\tau_1\gamma C^2(1+\sigma^2d)}{2\tau_2\mu}).
    \end{aligned}
\end{equation}
Our DP-SAD converges when we guarantee that $0<2\tau_2\gamma\mu<1$ and the error from the minimum $\mathcal{L}(\theta^*)$ is $\frac{\tau_1\gamma C^2(1+\sigma^2d)}{2\tau_2\mu}$.

\subsection*{Experimental Details}
\myPara{Datasets.}~MNIST and FashionMNIST are both 10-class datasets containing 60,000 training images and 10,000 testing images. Each image is 28$\times$28 grayscale image. CelebA is a face attribute dataset, which contains 202,599 color images of celebrity faces. We use the official preprocessed version with the face alignment and resize the images to 64$\times$64$\times$3. We create CelebA-H and CelebA-G based on it. CelebA-H is a classification dataset with hair color (black/blonde/brown) as the label and CelebA-G is a classification dataset with gender as the label.

\myPara{Baselines.}~DP-GAN is to directly apply the DPSGD training strategy to the training process of WGAN. Because WGAN itself satisfies the Lipschitz condition, the effect from gradient clipping is eliminated. PATE-GAN, DP-MERF, GS-WGAN, P3GM, G-PATE and DataLens are all based on PATE framework with different teacher aggregation strategies. All of the above baselines achieve differential privacy based on the Gaussian mechanism. DPGEN achieves differential privacy based on randomized response mechanism. PSG incorporates the downstream task into training to improve its data quality, but it requires repeated training for different downstream tasks. Both DP-DM and DP-LDM are implemented by applying DPSGD directly to diffusion models for differentially private generative modeling. We get the experimental results by running official codes or from original papers.

\begin{figure}[!htbp]
	\centering
 \includegraphics[width=0.7\linewidth]{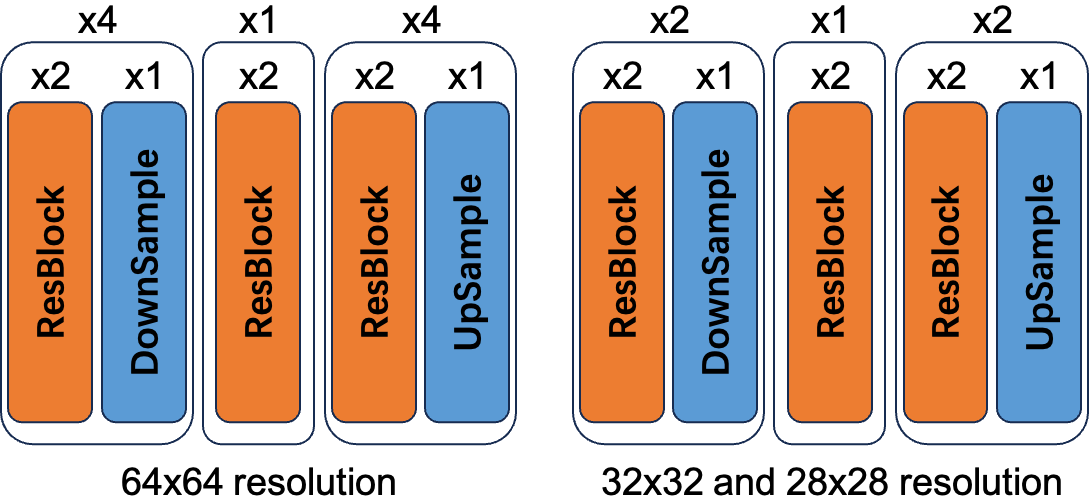}
\caption{Model structure for different resolutions.}\label{fig:network}
  \vspace{-10pt}
\end{figure}

\myPara{Implementations.}~We set the norm bound $C$ to $10^{-6}$. We set the training epoch to $100$ for all models and compute the $\sigma$ by RDP. We set the trade-off weight $\lambda$ and time step $T$ to $1$ and $500$, respectively. We set the initial values of both $\gamma$ and $\gamma_d$ to $10^{-4}$, and employ a ``CosineAnnealingLR'' to adjust them. We set $\omega$ in Eq.(5) to 1.8, batch size to 128, $\beta_{start}$ to 1e-4, $\beta_{end}$ to 0.028, $\sigma$=1.9 for $\varepsilon$=1 and $\sigma$=0.6 for $\varepsilon$=10. A simplified version of the model structures for different resolutions is shown in Fig.~\ref{fig:network}

\subsection*{Disscusion about different loss functions}

To investigate the impact of the loss function, we compare the classification accuracy on MNIST with different loss functions. The results are shown in Tab.~\ref{tab:1}. The adversarial loss (ADV) forces the student manifold to conform to the teacher manifold, which accelerates the convergence process. The teacher MSE loss (MSE-T) can also accelerate the convergence process. When the privacy budget~($\varepsilon$) is determined, faster convergence leads to a model with better performance.

\begin{table}[!htbp]
\caption{Classification accuracy comparisons on MNIST with different loss functions.}\label{tab:1}
\small
\begin{center}
\begin{tabular}{ccc|c|c|c}
    {MSE-T} & {MSE} & {ADV} & e=20 & e=50 & e=100 \cr
    \hline
    \checkmark & & & 0.5811 & 0.7452 & 0.9499 \cr
    \hline
    \checkmark & \checkmark && 0.6058 & 0.7617 & 0.9512 \cr
    \hline
    \checkmark & \checkmark & \checkmark & \textbf{0.6949} & \textbf{0.8272} & \textbf{0.9761} \cr
\end{tabular}\end{center}
\end{table}
\vspace{-8mm}

\subsection*{Limitations}
There are still some limitations to DP-SAD: i) because we need to set a large time step $T$ to dilute the effect of DP noise, which leads to low efficiency in sampling the generated images. Unlike GANs, which require only a single inference step to generate the final image, our method necessitates $T$ inference iterations to produce the final image; ii) our method does not employ the architecture of latent diffusion models; instead, it integrates with a VAE. The architecture of latent diffusion models is more advantageous for generating high-resolution images; iii) inevitably, the model might inadvertently assimilate the biases present within the dataset. In future research, efforts could be directed toward mitigating the acquisition of these biases by incorporating specific prompts during the model's training phase.

\section{Extened Visualization Results}
\begin{figure}[!htbp]
	\centering
	\includegraphics[width=0.45\linewidth]{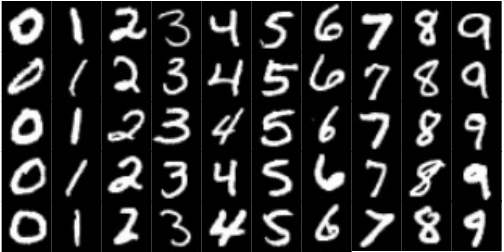}
        \includegraphics[width=0.45\linewidth]{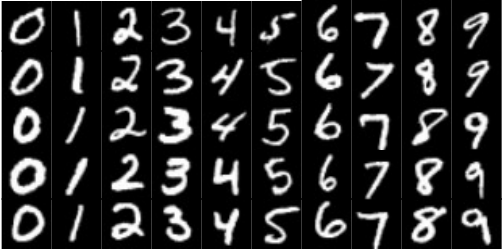}
	\caption{Visualization results of MNIST with 28$\times$28 resolution under $\varepsilon=1$ (left) and $\varepsilon=10$ (right).}
	\label{fig:mnist}
\end{figure}

\begin{figure}[!htbp]
	\centering
	\includegraphics[width=0.45\linewidth]{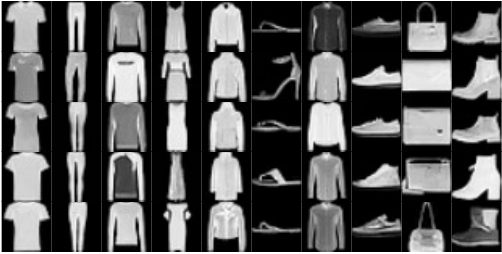}
        \includegraphics[width=0.45\linewidth]{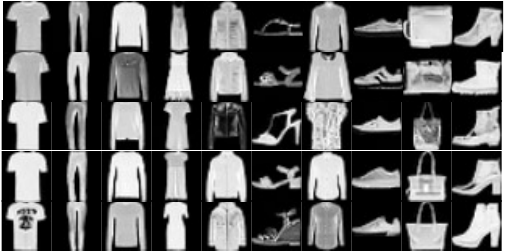}
	\caption{Visualization results of FMNIST with 28$\times$28 resolution under $\varepsilon=1$ (left) and $\varepsilon=10$ (right).}
	\label{fig:fmnist}
\end{figure}

\begin{figure}[!htbp]
	\centering
	\includegraphics[width=0.88\linewidth]{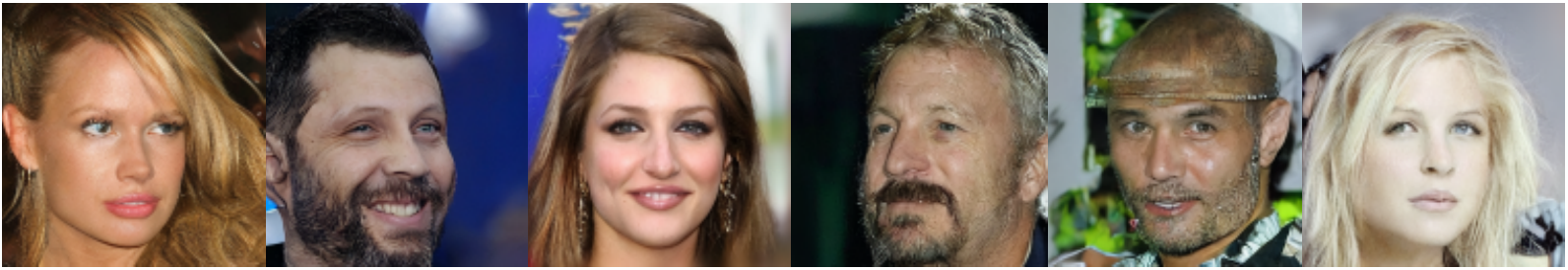}
	\caption{Visualization of CelebA at 128$\times$128 resolution.}
	\label{fig:vis}
\end{figure}

We further visualize the generated results on three datasets, including MNIST, FMNIST and CelebA, under different privacy budget. The results are shown in Fig.~\ref{fig:mnist}, Fig.~\ref{fig:fmnist}, Fig.~\ref{fig:celeba} and Fig.~\ref{fig:vis}. We find that there is not much difference visually between the results for $\varepsilon=1$ and $\varepsilon=10$, which demonstrates the potential of our method to generate higher-resolution images with stronger privacy protection.

\begin{figure}[!htbp]
	\centering
	\includegraphics[width=0.45\linewidth]{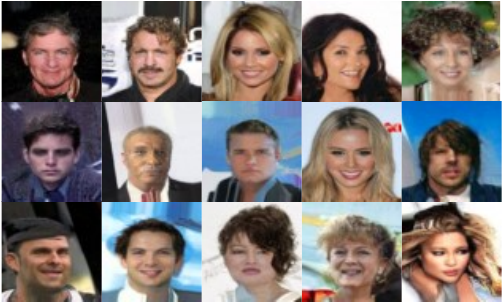}
        \includegraphics[width=0.45\linewidth]{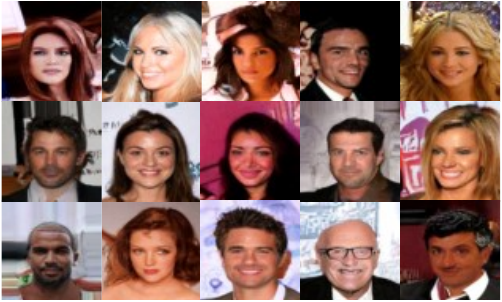}
	\caption{Visualization results of CelebA with 64$\times$64 resolution under $\varepsilon=1$ (left) and $\varepsilon=10$ (right).}
	\label{fig:celeba}
\end{figure}

\end{document}